\documentclass[letterpaper, 10 pt, conference]{ieeeconf}  
\makeatletter
\def\endtable{\end@float}
\makeatother

\IEEEoverridecommandlockouts                              

\usepackage{graphics} 
\usepackage{graphicx}
\usepackage{graphicx}
\usepackage{epsfig} 
\usepackage{mathptmx} 
\usepackage{times} 
\usepackage{amsmath} 
\usepackage{amssymb}  
\usepackage{amsmath}
\usepackage{booktabs}
\usepackage{float}
\usepackage{microtype}

\usepackage{amsmath,amsfonts}
\usepackage{algorithmic}
\usepackage{algorithm}
\usepackage{array}
\usepackage[caption=false,font=normalsize,labelfont=sf,textfont=sf]{subfig}
\usepackage{textcomp}
\usepackage{stfloats}
\usepackage{url}
\usepackage{verbatim}
\usepackage{graphicx}
\usepackage{cite}

\usepackage{booktabs} 
\usepackage{tabularx} 
\usepackage{makecell} 
\usepackage{adjustbox} 
\title{\LARGE \bf
Multi-Objective Trajectory Planning for a Robotic Arm in Curtain Wall Installation
}

\author{Xiao Liu$^{1}$, Yunxiao Cheng$^{1}$, Weijun Wang$^{1}$, Tianlun Huang$^{1}$, Zhiyong Wang$^{1}$,   Wei Feng $^{1,2*}$
\thanks{*corresponding author. e-mail: wei.feng@siat.ac.cn}
\thanks{$^{1}$ All authors are with Shenzhen Institute of Advanced Technology, Chinese Academy of Sciences, Shenzhen, 518055, China.
        {\tt\small  Contact: xiao.liu1@siat.ac.cn}}%
 \thanks{$^{2}$ All authors are with Shenzhen University of Advanced Technology and University of Chinese Academy of Sciences, Shenzhen, 518055, China.
        {\tt\small  Contact: wei.feng@siat.ac.cn}}%
}

\begin{document}

\maketitle
\thispagestyle{empty}
\pagestyle{empty}

\begin{abstract}
In the context of labor shortages and rising costs, construction robots are regarded as the key to revolutionizing traditional construction methods and improving efficiency and quality in the construction industry. In order to ensure that construction robots can perform tasks efficiently and accurately in complex construction environments, traditional single-objective trajectory optimization methods are difficult to meet the complex requirements of the changing construction environment. Therefore, we propose a multi-objective trajectory optimization for the robotic arm used in the curtain wall installation. First, we design a robotic arm for curtain wall installation, integrating serial, parallel, and folding arm elements, while considering its physical properties and motion characteristics. In addition, this paper proposes an NSGA-III-FO algorithm (NSGA-III with Focused Operator, NSGA-III-FO) that incorporates a focus operator screening mechanism to accelerate the convergence of the algorithm towards the Pareto front, thereby effectively balancing the multi-objective constraints of construction robots. The proposed algorithm is tested against NSGA-III, MOEA/D, and MSOPS-II in ten consecutive trials on the DTLZ3 and WFG3 test functions, showing significantly better convergence efficiency than the other algorithms. Finally, we conduct two sets of experiments on the designed robotic arm platform, which confirm the efficiency and practicality of the NSGA-III-FO algorithm in solving multi-objective trajectory planning problems for curtain wall installation tasks.
\end{abstract}

\section{INTRODUCTION}

In the current construction landscape, construction robots are an important way to improve construction efficiency and quality[1]-[2]. Trajectory planning can ensure that robots execute construction operations optimally, thereby minimizing construction time and energy consumption. Optimizing the motion trajectory of construction robots is crucial[3]–[4]. The optimization of motion trajectories primarily involves three aspects: time[5]–[7], energy consumption[8], and joint impact[9]–[10] during motion.

Multi-objective trajectory planning typically involves more complex tasks and constraints, and requires consideration of the priorities and resolution of conflicts between different objectives. Solving this problem has always been a popular topic in academic research[11]–[14]. [15] proposes a multi-objective trajectory optimization method based on response surface methodology (RSM) and non-dominated sorting genetic algorithm III (NSGA-III). [5] proposes a 3-5-3 polynomial interpolation trajectory planning algorithm based on an improved cuckoo search algorithm (ICS) that functions under a velocity constraint. The proposed algorithm performed well and realized a better time-optimal trajectory. [16] efficiently addressed the trajectory-planning problem for kiwifruit harvesting manipulators using multi-objective trajectory planning based on NSGA-III. [17] proposed an improved elitist non-dominated sorting genetic algorithm (INSGA-II) by introducing three genetic operators: ranking group selection (RGS), direction-based crossover (DBX), and adaptive precision-controllable mutation (APCM), which was developed to optimize travelling time and torque fluctuation.

Most of the existing research and methods have focused on industrial robotic arms. To address the challenges faced by construction robots, such as long working hours, high energy consumption, and complex tasks, we propose a multi-objective trajectory planning method for our designed robotic arm. This method introduces a focused operator selection mechanism and a new algorithm, NSGA-III-FO (NSGA-III with Focused Operator, NSGA-III-FO), to enhance the population selection process and accelerate convergence towards the Pareto front. The proposed algorithm demonstrates significant performance advantages in ﻿convergence efficiency, effectively balancing multiple objectives to achieve solution sets that meet task requirements.
 \begin{figure}[thpb]
      \centering
      \includegraphics[scale=0.1]{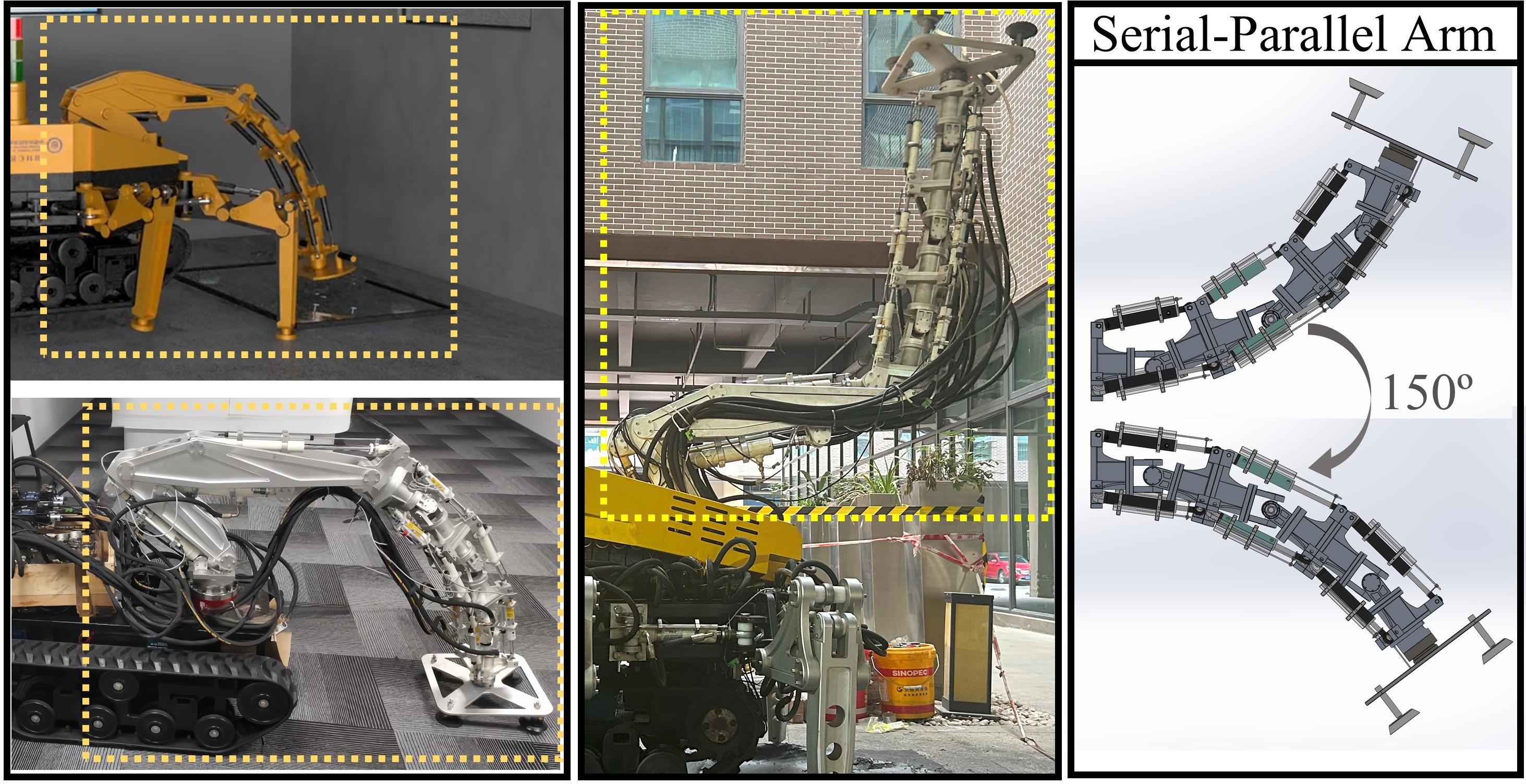}
      \caption{Structural Design of a Robotic Arm for Curtain Wall Installation}
      \label{figurelabel}
   \end{figure}
\section{DESIGN OF A ROBOTIC ARM FOR CURTAIN WALL INSTALLATION}
Currently, serial robots struggle to meet the high load-to-weight ratio requirements for construction tasks. Leveraging the high load-bearing capacity of parallel mechanisms with redundant actuation and the superior workspace characteristics of serial mechanisms, we design a robotic arm for curtain wall installation. The proposed arm includes a folding arm, a serial-parallel arm, and an end effector, as illustrated in Figure 1. The folding arm consists of a base platform, two support arms, three hydraulic cylinders, and an end platform. The serial-parallel arm comprises three 2-DOF parallel mechanisms connected in series, each containing three hydraulic cylinders(two active axes and one passive axis). All joints are driven by hydraulic cylinders. The entire robotic arm has six degrees of freedom and can perform compliant operations over a 150-degree range. The wrist motor enables ±360-degree rotation of the end effector. The end effector consists of four suction cups and a suction cup frame, controlled by a solenoid valve for overall air circuit switching, designed for glass handling.

\section{TRAJECTORY PLANNING}
\subsection{Sixth-Order B-Spline Interpolation Trajectory}
Mapping from Task Space to Joint Space: To execute predefined trajectory tasks, B-spline curves define the trajectory in task space. Key points $ P_n $ are determined based on the velocity requirements, where $ n $ denotes the sequence number of key points. Moving from the initial point $ P_0 $ to the final point $ P_n $ in task space, a series of key points in joint space can be obtained via inverse kinematics algorithms.
\begin{equation}
\label{deqn_ex1}
P_{in}=(q_{in},t_n),
\end{equation}
In the equations: $ i $ denotes the joint of the robotic arm; $ q_{in} $ represents the joint angle at key point $ n $ for joint $ i $, and $ t_n $ indicates the time at which the trajectory passes through key point $ P_{in} $. To construct a sixth-order B-spline curve based on the given key points, ensuring that each joint trajectory passes through these points $ p(x) $, B-spline interpolation is used to determine the control points that make the curve pass through these specified points.

To ensure that the robotic arm's joints pass through $ n+1 $ key points, we have: 
\begin{equation}
\label{deqn_ex1}
P\left(x_{i+6}\right)=\sum_{j=i}^{i+6}d_jN_{j,6}\left(x_{i+6}\right)
\end{equation}
Where: \( d_j \) denotes the control points; \( x \) is the knot vector, where \( x_{i+6} \in [x_6, x_{i+6}] \);  \( j \)isthe parameter of the B-spline curve; \( N_{j,6}(x) \) is the 6th-order normalized B-spline basis function, which depends on the vector \( x \) and is constructed recursively. The constraint function can be expressed as:
\begin{equation}
\label{deqn_ex1}
\begin{cases}
\dot{P}(x)|_{x_6}=\nu_s,\dot{P}(x)|_{x_6}=\nu_e \\
\ddot{P}(x)|_{x_6}=a_s,\ddot{P}(x)|_{x_6}=a_e \\
\dddot{P}(x)|_{x_6}=j_s,\dddot{P}(x)|_{x_6}=j_e & 
\end{cases}
\end{equation}
In the equations: \( v_s \) and \( v_e \) represent the initial and final velocities of the joint; \( a_s \) and \( v_e \) represent the initial and final accelerations of the joint; \( j_s \) and \( j_e \) represent the initial and final jerk of the joint. \( \dot{\mathbf{P}}(x) \), \( \ddot{\mathbf{P}}(x) \), and \( \dddot{\mathbf{P}}(x) \) denote the first, second, and third derivatives of the B-spline curve, respectively. The velocity, acceleration, and jerk for each joint in a sixth-order B-spline can be expressed as:
\begin{equation}
\label{deqn_ex1}
\begin{cases}
\nu(t)=\dot{P}(x)=\sum_{j=i-5}^id_j^1N_{j,5}(x) \\
a(t)=\ddot{P}(x)=\sum_{j=i-4}^id_j^2N_{j,4}(x) \\
j(t)=\dddot{P}(x)=\sum_{j=i-3}^id_j^3N_{j,3}(x) 
\end{cases}
\end{equation}
In the equations: \( \mathbf{d}_j^r = \left[ d_{1_j}^r, d_{2_j}^r, \ldots, d_{N_j}^r \right]^{\mathrm{T}} \) is the control point vector, and the derivative order \( r \) is 1, 2, or 3.

\subsection{Multi-Objective Trajectory Planning}
In real-world construction scenarios, energy consumption optimization is crucial for robots powered by batteries or generators. Joint impact optimization can reduce wear and tear on mechanical joints, thereby extending the robot lifespan. A high efficiency is also essential for the construction process. Therefore, the optimization objectives are time, joint impact, and energy consumption. Define the time evaluation function $ f_1 $, joint impact evaluation function $ f_2 $, and energy consumption evaluation function $ f_3 $ as follows: 
\begin{equation}
\label{deqn_ex1}
\begin{aligned}
 & \mathrm{f}=\sum_{i=0}^{n-1}\left(t_{i+1}-t_{i}\right) \\
 & f_{2}=\sum_{k=1}^{K}\sqrt{\frac{1}{T}\int_{0}^{T}\left(j_{i}\right)^{2}dt} \\
 & f_{3}=\sum_{k=1}^{K}\sqrt{\frac{1}{T}\int_{0}^{T}\left(\omega_{i}\tau_{i}\right)^{2}dt}
\end{aligned}
\end{equation}
In the equations:  \( K \)  and \( k \)  represent the total number of sampling points and the number of sampling points, respectively;  \( j_i \) is the angular jerk of joint \( i \); \( \omega_i \) is the angular velocity of joint \( i \);  \( \tau_i \)  is the output torque of joint \( i \). The motion constraints for the robotic arm are defined as follows: 
\begin{equation}
\label{deqn_ex1}
\begin{gathered}
\left|\tau_{i}(t)\right|\leq\tau_{i\max} \\
\left|j_{i}(t)\right|\leq j_{i\mathrm{max}} \\
\left|\omega_i(t)\right|\leq\omega_{i\max} \\
\left|\nu_{i}(t)\right|\leq\nu_{i\mathrm{max}}
\end{gathered}
\end{equation}
where: \(\tau_{i\max}\), \(j_{i\max}\), \(\omega_{i\max}\),\(v_{i\max}\) is the maximum torque, the maximum angular jerk, the maximum angular velocity, and the maximum linear velocity of joint $ i $.

\section{NSGA-III-FO MULTI-OBJECTIVE OPTIMIZATION ALGORITHM WITH FOCUS OPERATOR}
Researchers commonly employ heuristic multi-objective optimization algorithms to address robotic arm trajectory planning problems [18]–[19], aiming to find an optimal set of solutions that balance various performance metrics. NSGA-III is a multi-objective optimization algorithm [20] that utilizes a reference-point-based strategy to decompose the objective space, allowing each reference point to correspond with multiple solutions. However, NSGA-III overlooks the efficiency of converging towards the Pareto front. To enhance the convergence performance, we propose an enhanced version of NSGA-III named NSGA-III-FO(NSGA-III with Focused Operator, NSGA-III-FO). This algorithm introduces focused and non-focused operators to accelerate population screening, thereby improving the overall convergence efficiency.

\subsection{NSGA-III-FO Algorithm}

The fundamental idea of the NSGA-III-FO algorithm is to iteratively generate a set of non-dominated solutions, where each non-dominated solution approximates an optimal solution. In each iteration, the algorithm creates a new initial population based on the non-dominated solutions from the current solution set. The algorithm first selects individuals using focused operators for inclusion in the next generation, while directly excluding those selected by non-focused operators. The remaining individuals are then subjected to non-dominated sorting to form the next generation solution set. This process continuously optimizes multiple objective functions simultaneously, ultimately yielding an optimal set of target allocation schemes. The optimization flowchart is illustrated in Figure 2.

\begin{figure}[htb]
      \centering
      \includegraphics[width=\columnwidth]{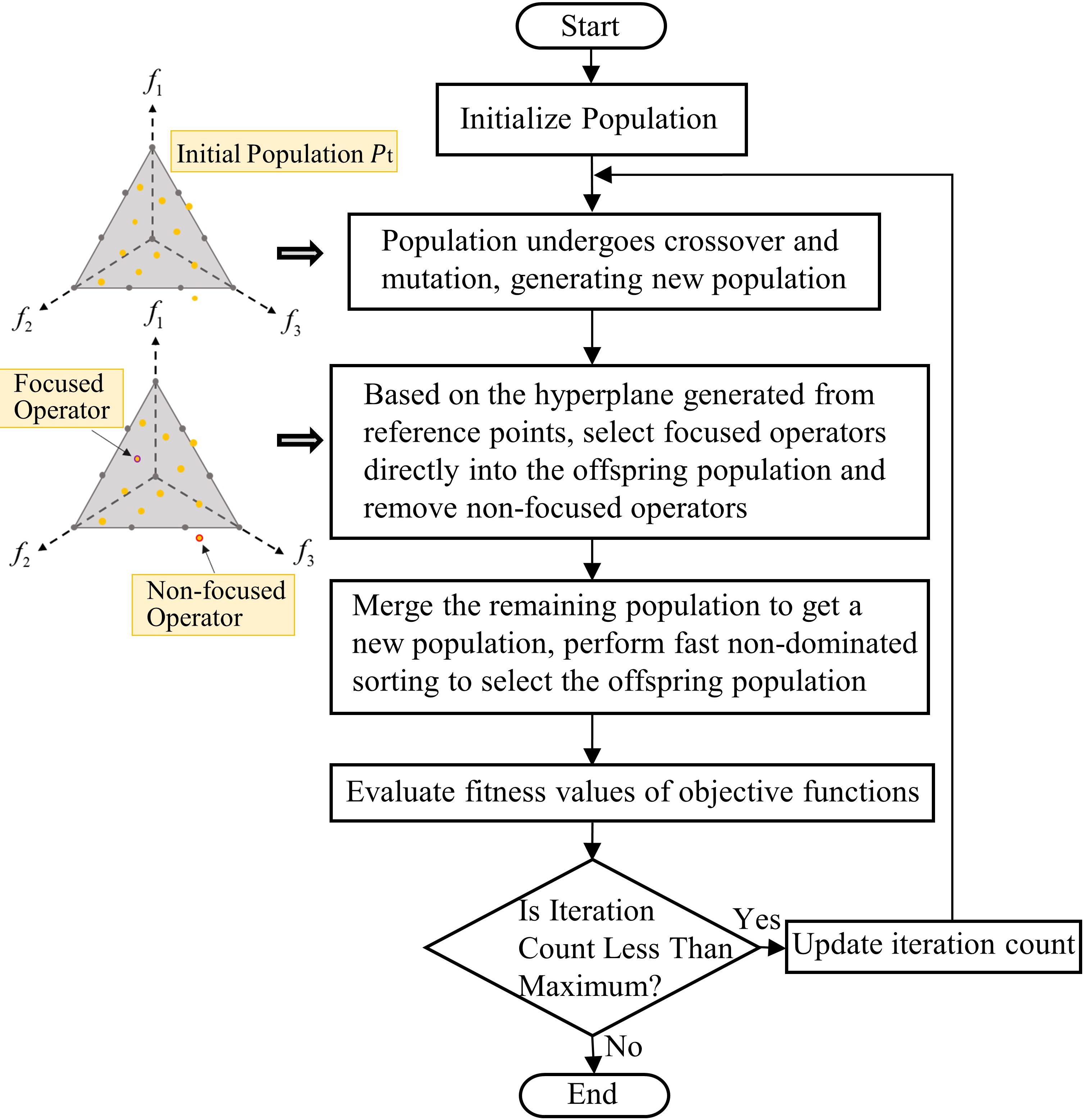}
      \caption{Flowchart of NSGA-III-FO for Multi-objective Optimization)}
      \label{figurelabel}
   \end{figure}
   
The specific solution steps of the NSGA-III-FO algorithm are as follows:

(1) Randomly initialize an initial population \( P_t \) of size N, construct a reference plane based on boundary crossover, for m objective optimization functions, divide into p on the m-dimensional standardized hyperplane, and uniformly generate H reference points. The calculation method is as follows:
\begin{equation}
\label{deqn_ex1}
H=\frac{(m+p-1)!}{p!(m-1)!}\quad 
\end{equation}

(2) In the initial population \( P_{t} \), select the individual with the smallest Euclidean distance from the reference plane as the focused operator, which directly enters the offspring population, and select the individual with the largest Euclidean distance from the reference plane as the non-focused operator to be directly excluded.

(3) The remaining individuals undergo non-dominated sorting, and the better individuals are selected from the sorting results to undergo crossover and mutation operations. The crossover probability and mutation probability are calculated as follows:
\begin{equation}
\label{deqn_ex1}
P_c =
\begin{cases}
P_{c,\max} - \frac{P_{c,\max} - P_{c,\min}}{1 + e^{\cos\left[ \left(\frac{\bar{f} - f_{\max}}{\bar{f} - f_{\min}}\right) \pi \right]}}, & f_{\max} \leq \bar{f} \\
P_{c,\max}, & f_{\max} > \bar{f}
\end{cases}
\end{equation}

\begin{equation}
\label{deqn_ex1}
P_m =
\begin{cases}
P_{m,\max} - \frac{P_{m,\max} - P_{m,\min}}{1 + e^{\cos\left[ \left(\frac{\bar{f} - f_{\max}}{\bar{f} - f_{\min}}\right) \pi \right]}}, & f_{\max} \leq \bar{f} \\
P_{m,\max}, & f_{\max} > \bar{f}
\end{cases}
\end{equation}

In the formula, $P_{c.max}$ and $P_{c.min}$ represent the maximum and minimum values of the set crossover probability, respectively, while $P_{m.max}$ and $P_{m.min}$ represent the maximum and minimum values of the set mutation probability, respectively. $\overline{f}$ is the average fitness function value of the current solution, and $f_{max}$ and $f_{min}$ are the maximum and minimum values of the fitness function for the current solution, respectively. After crossover and mutation produce offspring populations $Q_t$, both populations are combined to form a new offspring population $R_t$ of size $2N$.

(4) Perform rapid non-domination sorting, retaining excellent and diverse individuals according to the non-domination relationship, forming a new parent population $P_{t+1}$.

(5) Generate a new offspring population $Q_{t+1}$ through basic operations of genetic algorithms, merge $P_{t+1}$ with $Q_{t+1}$ to form a new population $R$, repeat these operations until reaching the designated number of generations.

\subsection{Performance Testing of the NSGA-III-FO Algorithm}
The NSGA-III-FO algorithm is used to conduct ten comparative tests against the NSGA-III, MOEA/D, and MSOPS-II algorithms on the DTLZ3 and WFG3 test functions. DTLZ3 and WFG3, as classic benchmark functions in multi-objective optimization, are widely used to evaluate the algorithm performance. Their standardized status ensures fair assessment. Using these recognized benchmarks facilitates an effective comparison with the existing algorithms. An important performance evaluation metric in multi-objective optimization problems is the Inverse Generational Distance (IGD). This metric measures the distance between the non-dominated solution set generated by the algorithm and the reference non-dominated solution set, thereby evaluating the performance of the algorithm. The smaller the IGD value, the closer the non-dominated solution set generated by the algorithm is to the reference non-dominated solution set, indicating better algorithm performance. The calculation method is:
\begin{equation}
\label{deqn_ex1}
 IGD\left(P,P^{*}\right) = \frac{\sum_{i=1}^{|P|}d\left(P_{i},P^{*}\right)}{|P^{*}|} 
\end{equation}

In the formula, $P$ represents the non-dominated point set in the target space; $P^*$ represents the uniformly distributed points on the true Pareto frontier; $|P|$ represents the number of solutions in set $P$; $d\left(P_i, P^*\right)$ represents the Euclidean distance between the solution $P_i$ and $P^*$ in the target space. The calculation result is shown in the table below:
\begin{table}[ht]
    \centering
    \caption{Comparison of IGD Values Among NSGA-III-FO and Other Multi-objective Algorithms}
    \resizebox{8.5cm}{!}{%
    \begin{tabular}{ccccc}
        \toprule
        Test Function & NSGA-III & MSOPS-II & MOEA/D & NSGA-III-FO \\
        \midrule
        DTLZ3 & 343.9±24.9 & 352.4±28.0 & 387.1±23.8 & 341.3±25.3  \\
        WFG3 & 0.6122±0.038 & 0.7547±0.048 & 0.6932±0.051 & 0.6087±0.028  \\
        \bottomrule
    \end{tabular}
    }
\end{table}

The hypervolume(HV)is an important metric for evaluating the quality of Pareto solution sets in multi-objective optimization algorithms. It measures the volume enclosed by the Pareto front and a reference point. Hypervolume can simultaneously reflect the convergence and diversity of the solution set. The hypervolume convergence curves for the DTLZ3 and WFG3 test functions, using the origin as the reference point in the Pareto solution space, are shown in the following figure:
\begin{figure}[htb]
      \centering
      \includegraphics[width=\columnwidth]{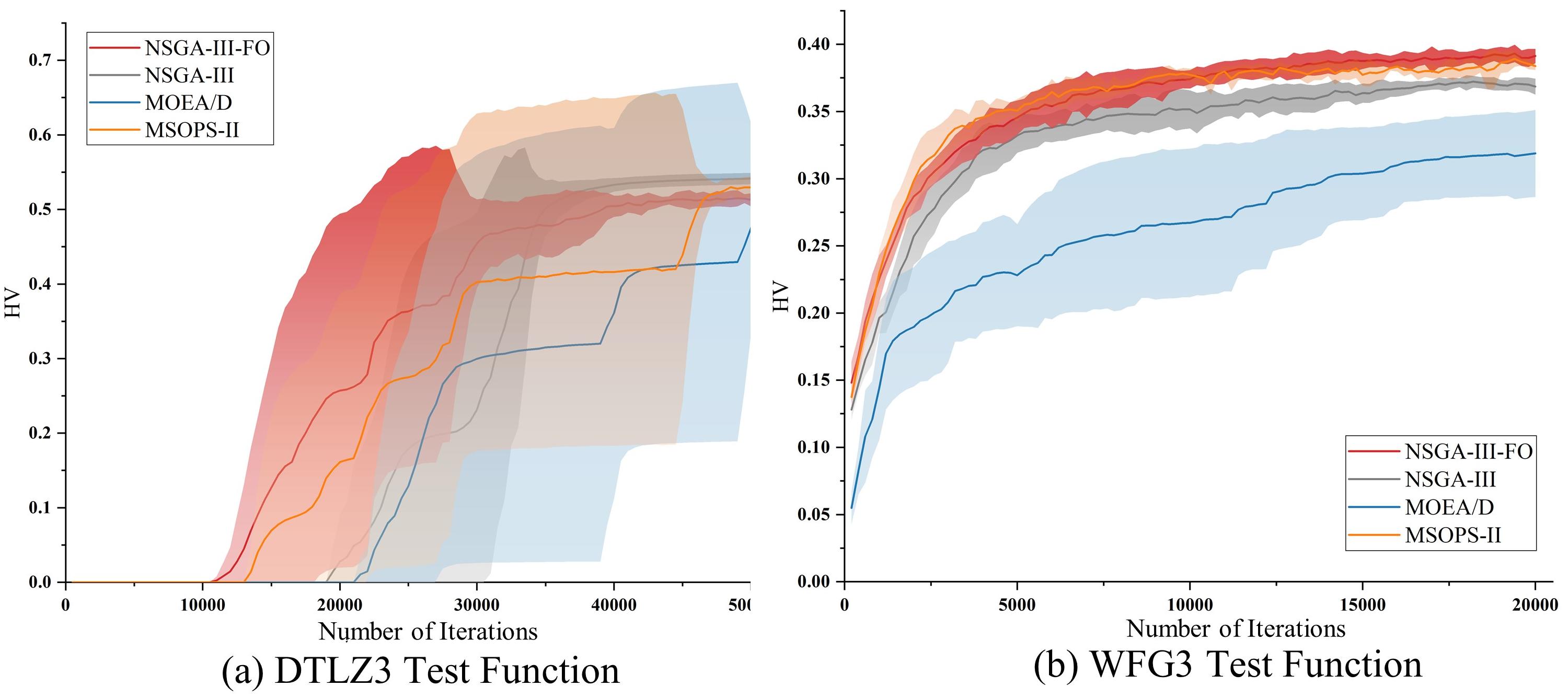}
      \caption{The HV Convergence Curves of Each Algorithm (Shaded Areas Indicate Standard Deviation Ranges)}
      \label{figurelabel}
   \end{figure}
From Figure 3 and Table 2, it can be observed that when the number of iterations is 20,000, the HV value of MSOPS-II for the DTLZ3 test function is only 63.6\% of that of NSGA-III-FO, while other algorithms have not yet found effective solutions. When the number of iterations reaches 50,000, the HV values of NSGA-III-FO, NSGA-III, and MSOPS-II tend to converge and their results are close, while the curve of MOEA/D has not yet converged. By analyzing the calculation results of IGD values, it is evident that NSGA-III-FO exhibits better convergence and stability.

In the WFG3 test function, at 10,000 iterations, the HV values of NSGA-III-FO, NSGA-III, and MSOPS-II have approached convergence, with NSGA-III-FO exhibiting the smallest standard deviation and stable convergence. In terms of IGD values, NSGA-III-FO outperforms the other three algorithms, with significantly lower average values. Although NSGA-III also performs well on WFG3, it still falls short compared to NSGA-III-FO.

\section{MULTI-OBJECTIVE TRAJECTORY PLANNING SIMULATION BASED ON NSGA-III-FO ALGORITHM}

We use the number of iterations as a convergence metric, where convergence speed reflects algorithm efficiency rather than absolute runtime. In task space, trajectories are defined using B-spline curves, and key points in the sequence of poses are determined based on trajectory task requirements. We set seven key points during the motion process, as shown in Table 2.

\begin{table}[ht]
    \centering
    \caption{Setting Key Points}
    \begin{tabular}{ccccccc}
        \toprule
        Node & Joint 1 & Joint 2 & Joint 3 & Joint 4 & Joint 5 & Joint 6 \\
        \midrule
        1 & 43.35 & 78.54 & -90.05 & 0 & 0 & 0 \\
        2 & 46.35 & 86.43 & -56.68 & 1.68 & 1.31 & 0.68 \\
        3 & 55.04 & 99.62 & -39.25 & 4.71 & 3.65 & 2.71 \\
        4 & 62.67 & 104.06 & -21.94 & 6.51 & 5.53 & 4.64 \\
        5 & 68.04 & 112.40 & -9.04 & 8.14 & 6.99 & 6.53 \\
        6 & 74.40 & 124.5 & 1.68 & 12.8 & 8.18 & 7.31 \\
        7 & 84.13 & 133.6 & 12.81 & 16.14 & 10.15 & 9.21 \\
        \bottomrule
    \end{tabular}
\end{table}
 Based on the dynamic parameters of a foldable serial-parallel robotic arm, we use the NSGA-III-FO algorithm to optimize the time function \(f_1\), impact function \(f_2\), and energy consumption function \(f_3\) through simulation. The multi-objective optimization trajectory Pareto solution set was obtained, as shown in Figure 4:
\begin{figure}[htb]
      \centering
      \includegraphics[width=\columnwidth]{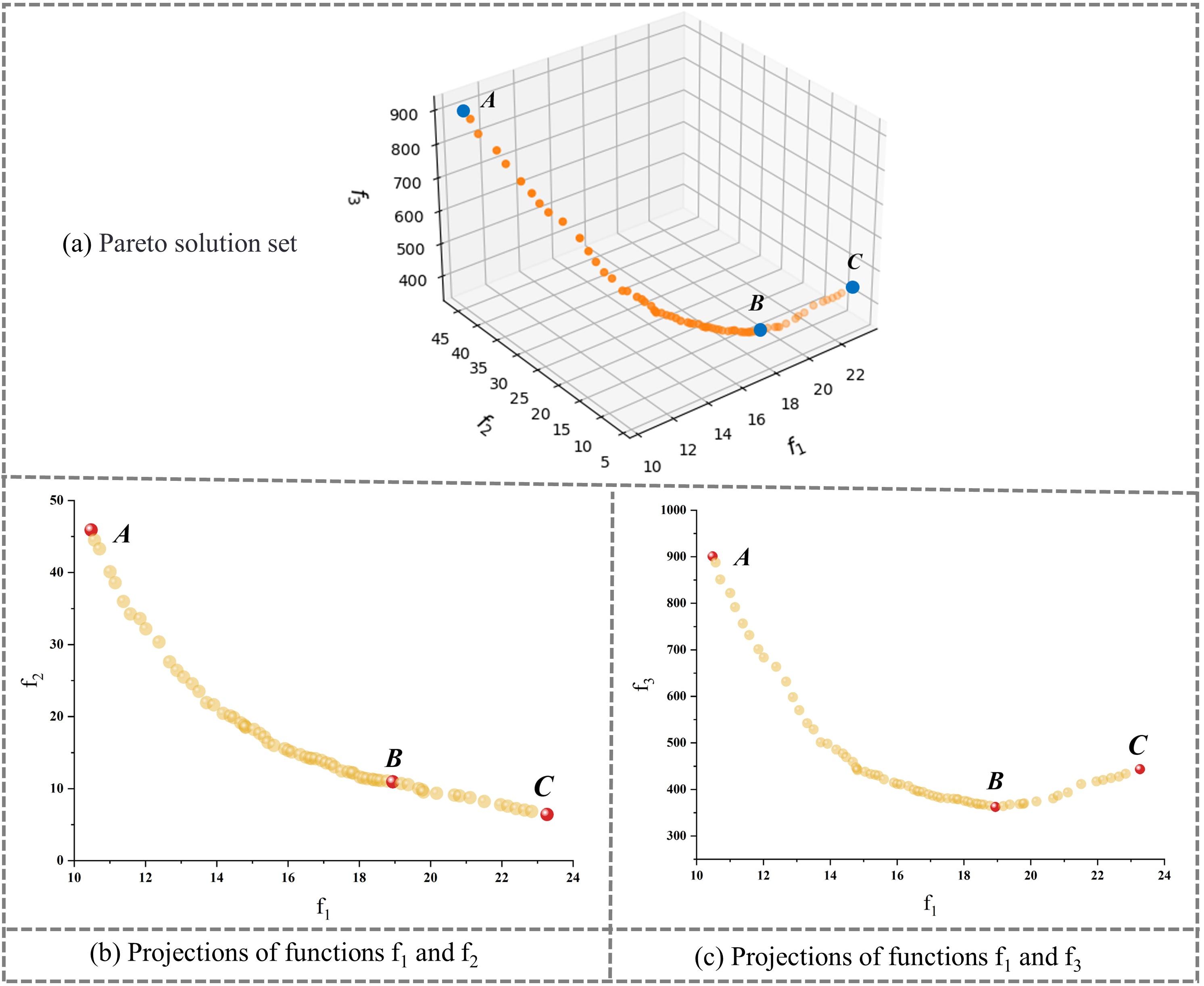}
      \caption{Pareto Solution Set of Multi-objective Optimized Trajectories Based on NSGA-III-FO Algorithm}
      \label{figurelabel}
   \end{figure}

As shown in Figure 4, A, B, and C represent the optimal solutions for \( f_1 \), \( f_2 \), and \( f_3 \), respectively. It can be observed that \( f_1 \) and \( f_2 \) exhibit a negative correlation, while the relationship between \( f_1 \) and \( f_3 \) is more complex but generally also shows a negative correlation. The time vectors of the optimal solutions for the objective functions \( f_1 \), \( f_2 \), and \( f_3 \), along with their corresponding function values, are shown in Table 3. The time vectors are the collection of time points corresponding to the seven key points in the motion process.

\begin{table}[ht]
    \centering
    \caption{Time Vectors and Corresponding Objective Function Values}
    \resizebox{9cm}{!}{%
    \begin{tabular}{ccccc}
        \toprule
        Plan & Time Vector(s)  &  $f_1$ (s)  & $f_2$ (N) & $f_3$ (J) \\
        \midrule
       A & [0, 2.01, 3.77, 5.79, 6.53, 8.85, 10.48] & 10.48 & 900.87 & 45.89 \\
       B & [0, 3.51, 7.34, 10.91, 13.31, 15.32, 18.94] & 18.94 & 362.25 & 10.95 \\
       C & [0, 1.51, 5.09, 10.23, 15.67, 18.98, 23.27] & 23.27 & 443.33 & 6.40 \\
        \bottomrule
    \end{tabular}
    }
\end{table}

The Pareto front of the solution set is observed to exhibit uniform distribution and good diversity. When applying the robotic arm to practical work requirements, different weights can be set for time, energy consumption, and joint impact metrics, allowing the selection of a multi-objective optimal solution that best meets the requirements. Additionally, the NSGA-III, MOEA/D, and NSGA-II algorithms are used for trajectory planning of the robotic arm with \( f_1 \), \( f_2 \), and \( f_3 \) as the objective functions, and compared with the NSGA-III-FO algorithm in terms of the HV metric. The convergence curves are shown in Figure 5.

\begin{figure}[thpb]
      \centering
      \includegraphics[width=\columnwidth]{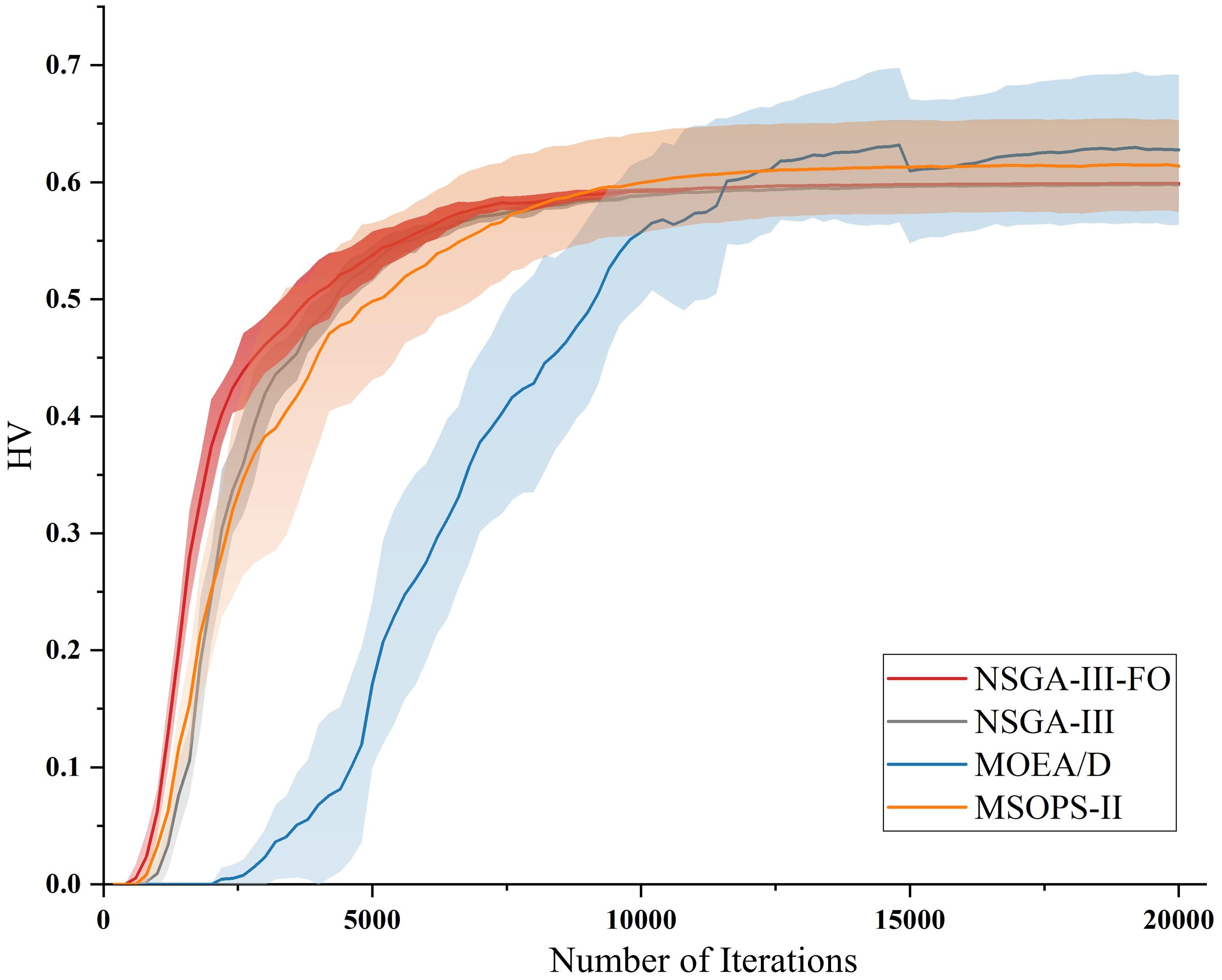}
      \caption{HV Indicator of Multi-objective Trajectory Planning Solutions for the robotic arm}
      \label{figurelabel}
   \end{figure}

The experimental results show that the NSGA-III-FO algorithm demonstrates a clear advantage in convergence rate, further validating its potential and effectiveness in exploring the solution space.
\section{EXPERIMENT ON MULTI-OBJECTIVE TRAJECTORY PLANNING}
\subsection{Robotic Arm Experimental Platform}
The curtain wall installation robotic arm we developed primarily consists of a folding arm, a serial-parallel manipulator, and an end effector. The folding arm is configured with three joints: Joint 1, Joint 2, and Joint 3. The serial-parallel manipulator is restricted to three degrees of freedom in the vertical plane, designated as Joint 4, Joint 5, and Joint 6. The experimental platform for the curtain wall installation robotic arm.is shown in Figure 6.

\begin{figure}[thpb]
      \centering
      \includegraphics[width=\columnwidth]{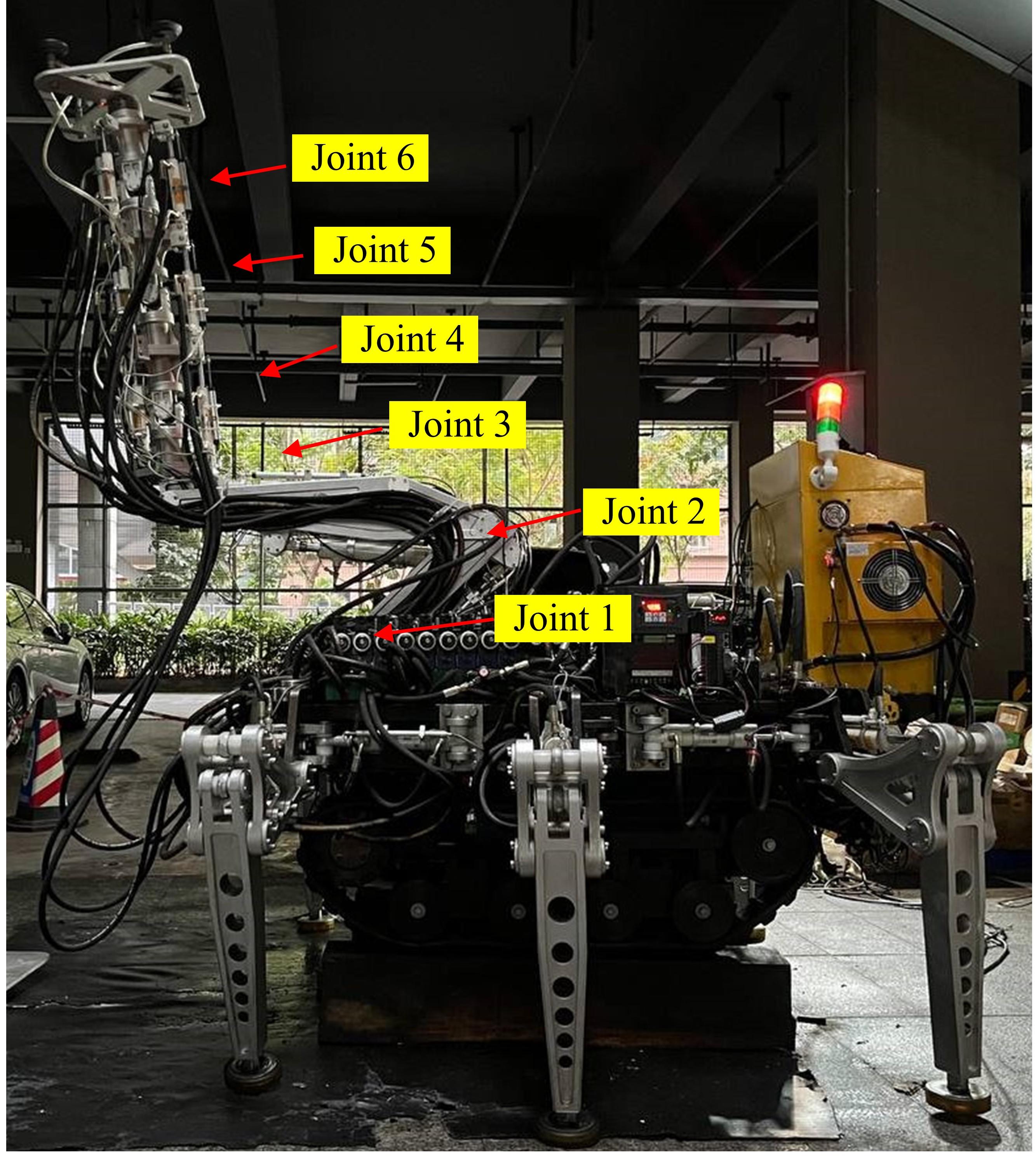}
      \caption{Robotic Arm Experimental Platform for Curtain Wall Installation Trajectory Planning}
      \label{figurelabel}
   \end{figure}

\subsection{Experiment}

We designed two construction tasks based on real-world scenarios to perform trajectory planning for the robotic arm. Utilizing the NSGA-III-FO algorithm, we successfully obtained solution sets that best met the task requirements. The feasibility of these solution sets was verified through experiments.

(1)Construction Task 1—Vertical Surface Installation: The task space is obstacle-free. Starting from the home position, the robotic arm picks up the steel plate and then transports it to the target position on a vertical surface directly in front of the arm. The steel plate weighs 10 kg and is picked up using four suction cups in 2 seconds. The task requirements are that \( f_1 \) must be less than 40, \( f_2\) must be less than 1000, and \( f_3 \) is not specified. Key points are detailed in Table 4.

\begin{table}[ht]
    \centering
    \caption{Joint Angles for Key Points in Task 1 (\textdegree)}
\label{tab:joint_angles}
\begin{tabular}{@{}ccccccc@{}}
\toprule
\textbf{Key Point} & \textbf{Joint 1} & \textbf{Joint 2} & \textbf{Joint 3} & \textbf{Joint 4} & \textbf{Joint 5} & \textbf{Joint 6} \\ \midrule
Key Point 1        & 30.0            & 130.0           & -60.0            & 0                & 0                & 0                \\
Key Point 2        & 37.2            & 126.4           & -16.8            & 6.8              & 8.8              & 10.4             \\
Key Point 3        & 73.2            & 119.2           & 19.2             & 14.7             & 15.8             & 16.0             \\
Key Point 4        & 80.0            & 104.0           & 19.0             & 16.0             & 18.0             & 18.0             \\
Key Point 5        & 80.0            & 104.0           & 19.0             & 16.0             & 18.0             & 18.0             \\
Key Point 6        & 61.2            & 108.4           & -1.6             & 12.8             & 11.8             & 10.8             \\
Key Point 7        & 57.2            & 103.6           & -12.8            & 10.0             & 10.0             & 10.0             \\ \bottomrule
\end{tabular}
\end{table}

The robotic arm's poses at key points during the experiment are shown in Figure 7.

\begin{figure}[thpb]
      \centering
      \includegraphics[width=\columnwidth]{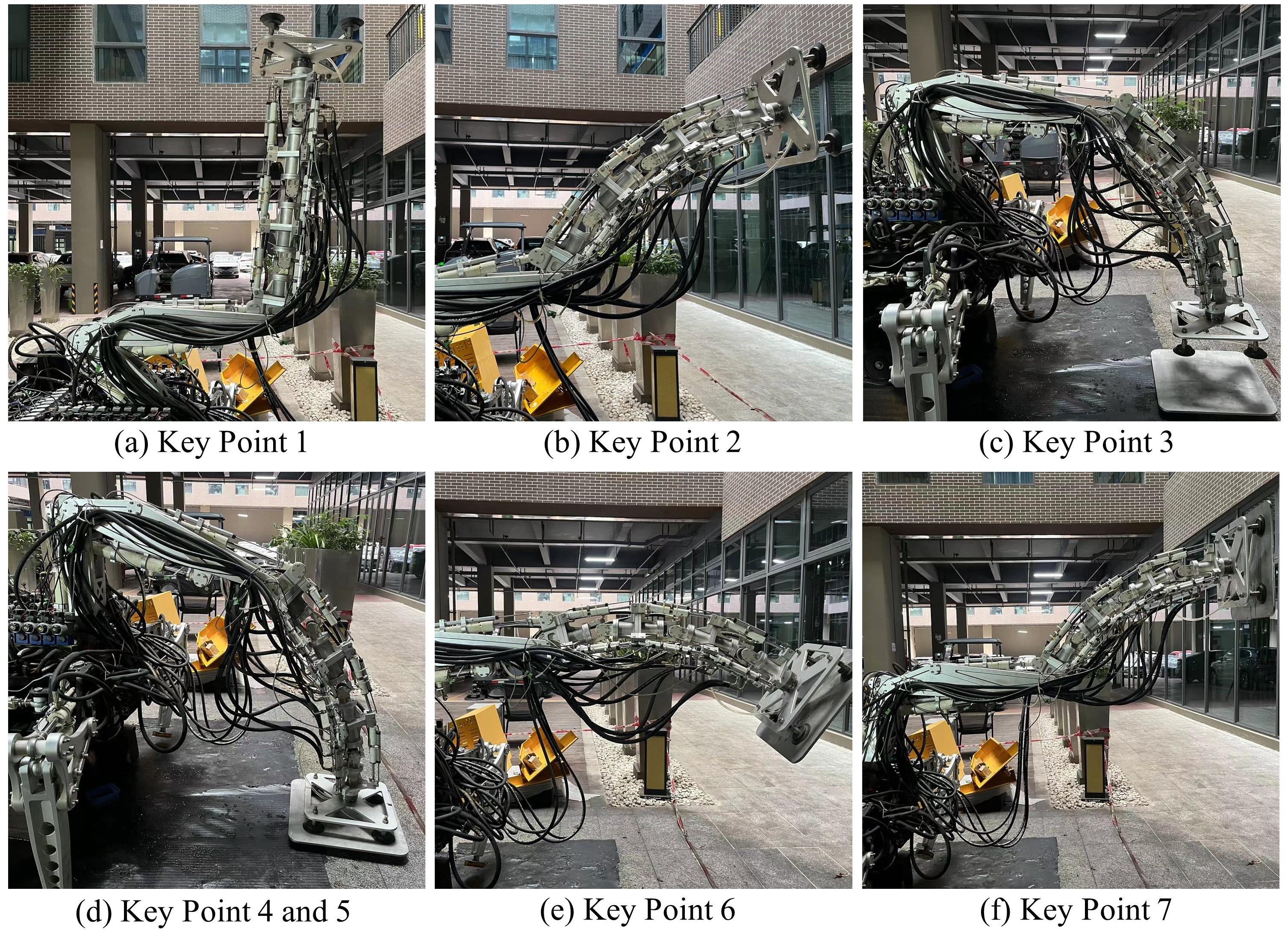}
      \caption{Robotic arm poses at key points during Task 1.}
      \label{figurelabel}
   \end{figure}

According to the key point requirements of Task 1, we first set the robotic arm to pass through key points 4 and 5 in 2 seconds. Next, we used the NSGA-III-FO algorithm to obtain the Pareto solution set. Finally, the trajectory was executed on the robotic arm platform. The time-optimal solution vector obtained from the NSGA-III-FO algorithm is [0, 7.744, 15.307, 22.863, 30.976, 38.720].

During Task 1 experiments, the changes in joint angles of the robotic arm and the measured joint angles are compared in Figure 8.
\begin{figure}[thpb]
      \centering
      \includegraphics[width=\columnwidth]{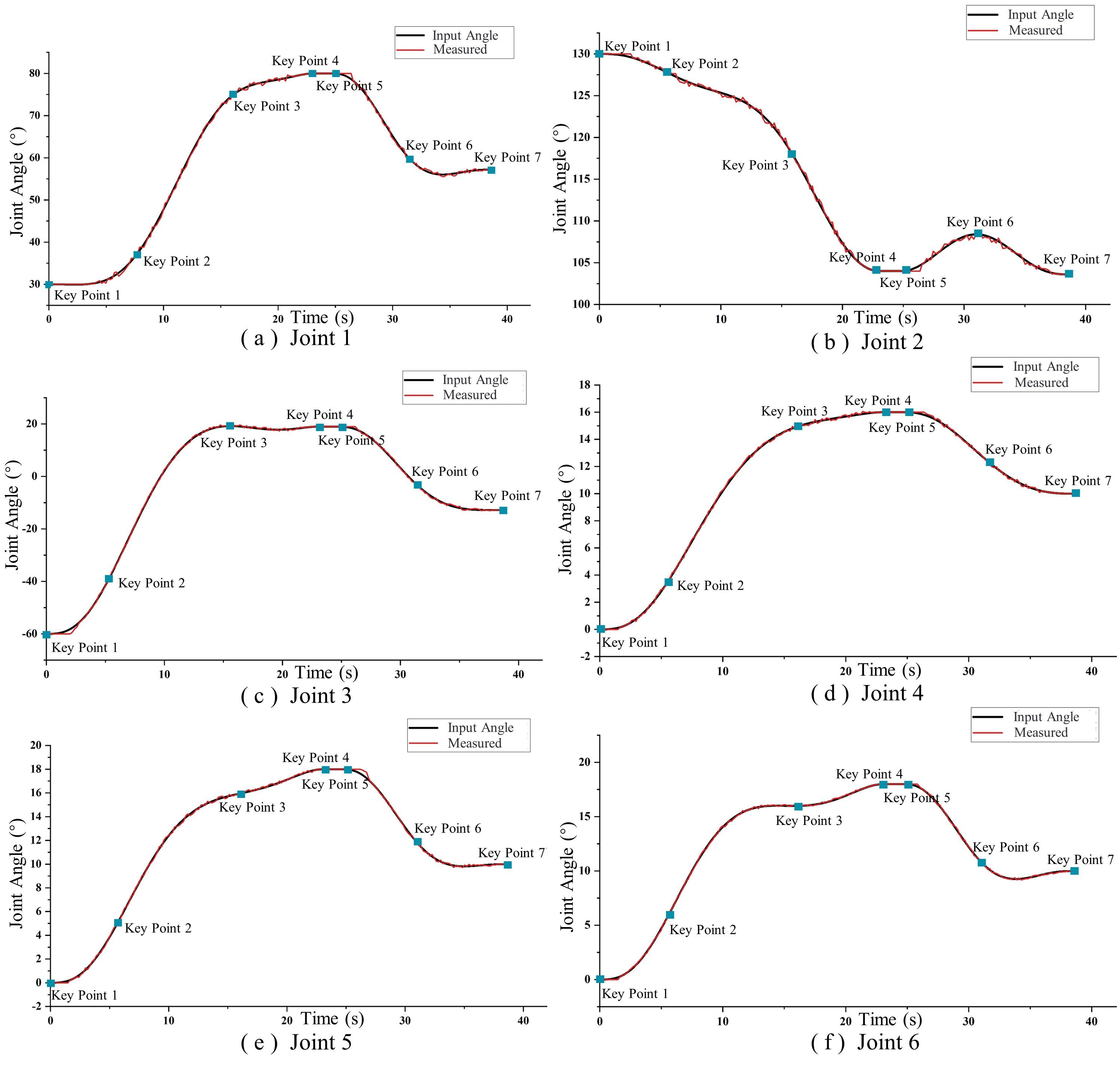}
      \caption{Joint Angle Variations of Robotic Arm for Task 1}
      \label{figurelabel}
   \end{figure}

In trajectory planning experiments, significant errors occurred during joint transitions from static to dynamic states, primarily due to hydraulic actuator precision limits and assembly inaccuracies. The current control algorithm does not effectively compensate for these errors. However, the robotic arm maintains stability, successfully passes through predefined key points, and completes trajectory tasks. Despite limitations in handling transition errors, the overall performance meets the basic requirements of trajectory planning.

(2) Construction Task 2—Overhead Panel Installation: In an obstacle-free space, the 
Starting from the home position, the robotic arm picks up a 10 kg steel plate and installs it overhead. The suction cup takes 2 seconds to pick up the steel plate. 
The task requirements are that \( f_1 \) must be less than or equal to 50, \( f_2 \)  must be less than 300, and \( f_3\) must be less than 65,000. Key points are detailed in Table 5.

\begin{table}[ht]
    \centering
    \caption{ Joint Angles for Key Points in Task 2 (\textdegree)}
\label{tab:joint_angles}
\begin{tabular}{@{}ccccccc@{}}
\toprule
  \textbf{Key Point} & \textbf{Joint 1} & \textbf{Joint 2} & \textbf{Joint 3} & \textbf{Joint 4} & \textbf{Joint 5} & \textbf{Joint 6} \\
        \midrule
        Key Point 1 & 30.0 & 130.0 & -60.0 & 0 & 0 & 0 \\
        Key Point 2 & 44.4 & 122.8 & -20.4 & 15.0 & 14.4 & 10.8 \\
        Key Point 3 & 80.0 & 104.0 & 19.0 & 16.0 & 18.0 & 18.0 \\
        Key Point 4 & 80.0 & 104.0 & 19.0 & 16.0 & 18.0 & 18.0 \\
        Key Point 5 & 51.6 & 108.4 & -24.0 & 14.0 & 13.4 & 10.4 \\
        Key Point 6 & 65.6 & 86.8 & -45.6 & 11.4 & 7.2 & 6.8 \\
        Key Point 7 & 73.0 & 57.8 & -47.2 & 0 & 0 & 0 \\
        \bottomrule
\end{tabular}
\end{table}

During the Task 2 experiment, the poses of the robotic arm at each key point are shown in Figure 9.

\begin{figure}[thpb]
      \centering
      \includegraphics[width=\columnwidth]{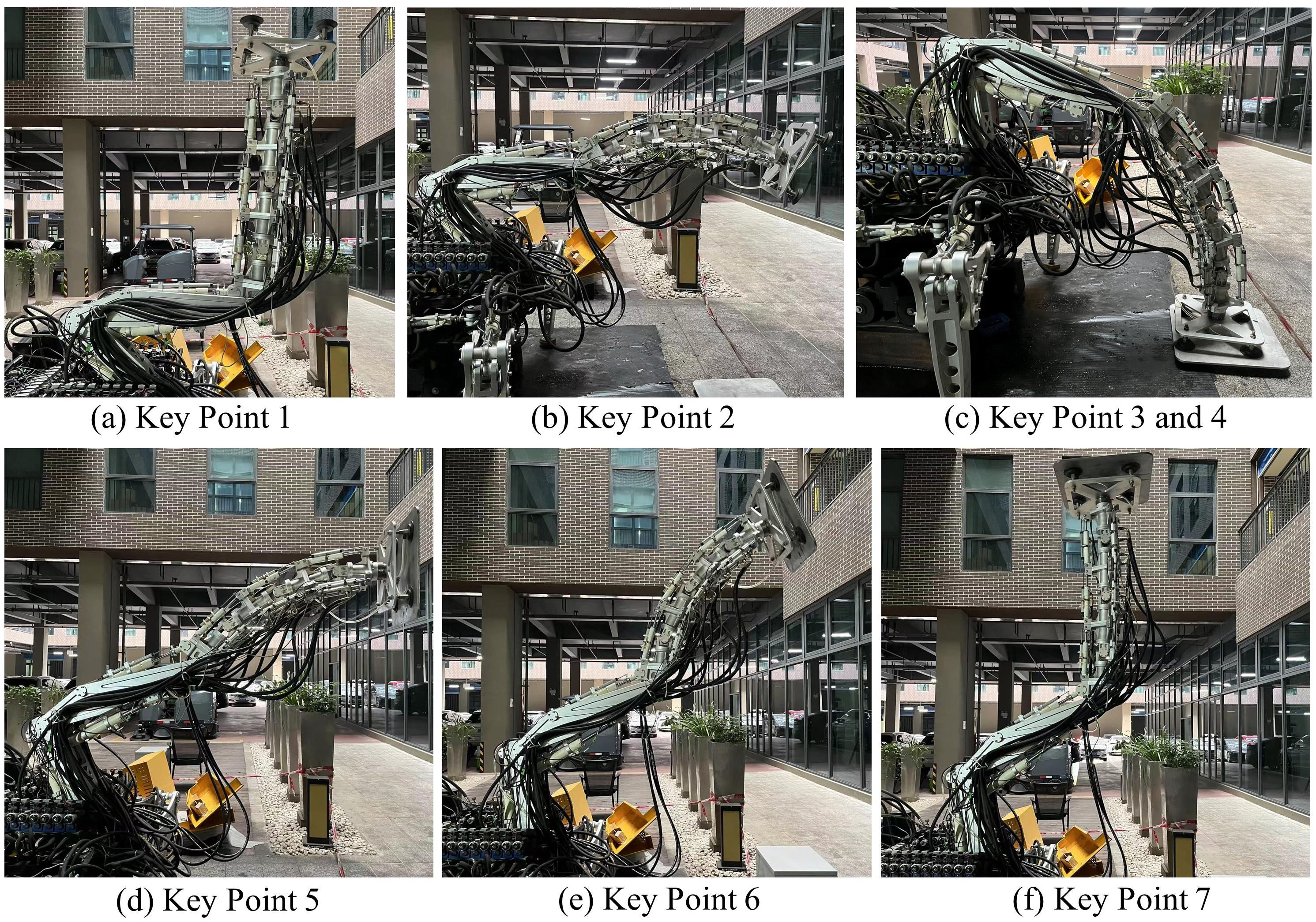}
      \caption{Robotic arm poses at key points during Task 2}
      \label{figurelabel}
   \end{figure}

According to the key point requirements of Task 2, we first set the robotic arm to pass through key points 3 and 4 in 2 seconds. We then used the NSGA-III-FO algorithm to obtain the Pareto solution set. Finally, the trajectory was executed on the robotic arm experimental platform. The multi-objective optimal time vector obtained from the NSGA-III-FO algorithm is [0, 9.431, 19.334, 21.456, 34.660, 39.140, 49.043].

During Task 2, the robotic arm required more time to complete the task, resulting in reduced joint angular velocities compared to previous tasks. This adjustment significantly decreased the observed errors during transitions from static to dynamic states. The robotic arm maintained accurate adherence to the predefined trajectory, successfully passing through all key points and completing the trajectory planning task.

In Task 2 experiments, the changes in joint angles of the robotic arm and the measured angles are shown in Figure 10.

\begin{figure}[thpb]
      \centering
      \includegraphics[width=\columnwidth]{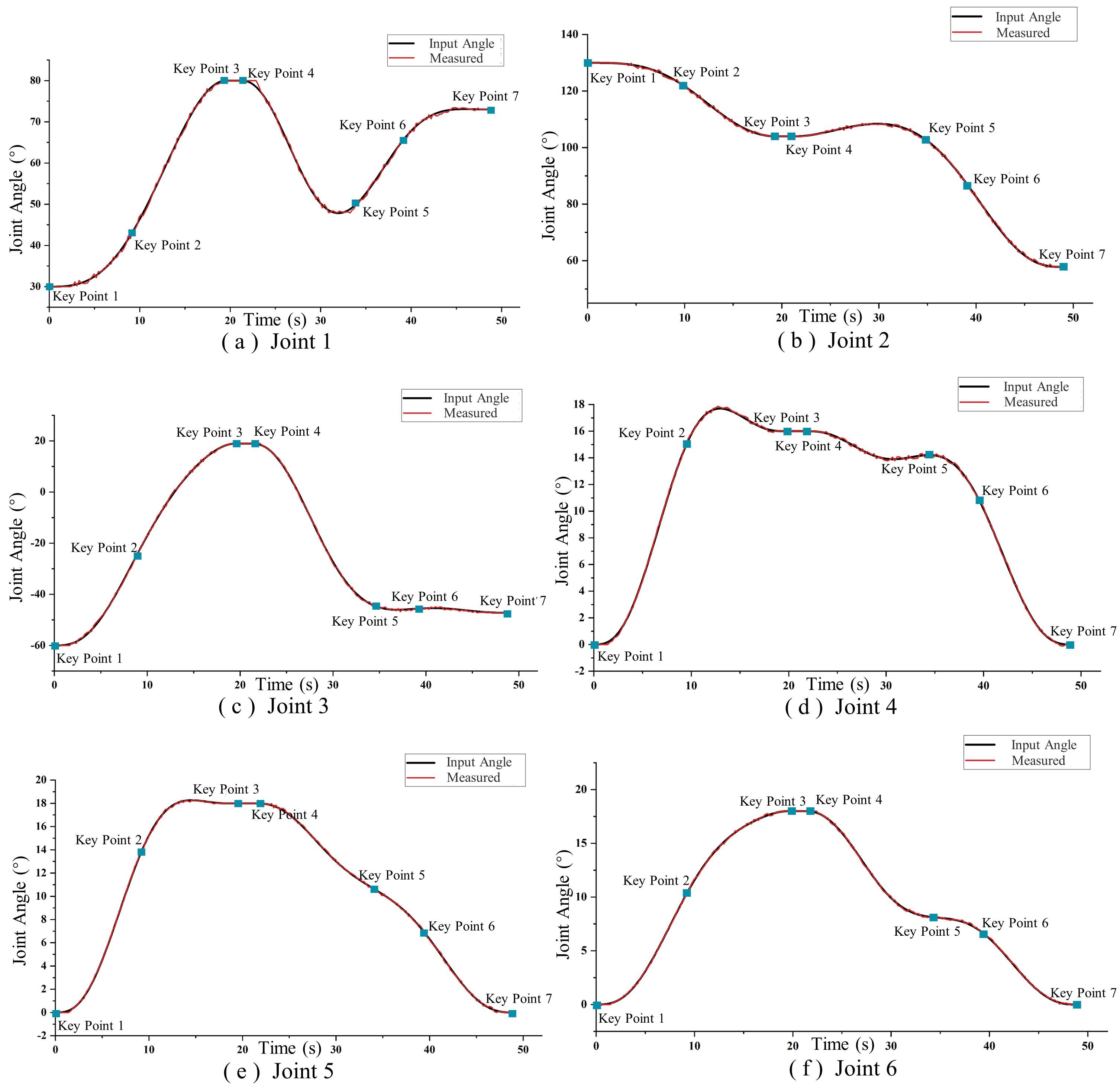}
      \caption{Joint Angle Variations of Robotic Arm for Task 2}
      \label{figurelabel}
   \end{figure}

Combining the results of the two task experiments, the robotic arm's end-effector actual trajectory closely matched the pre-simulated trajectory while satisfying all joint torque constraints. Additionally, the joint angle curves during actual motion were smooth, with no noticeable oscillations observed. Smooth transitions were achieved at all key points. This verifies the effectiveness and practicality of the NSGA-III-FO algorithm in multi-objective trajectory planning.

\section{CONCLUSIONS}

In this study, we developed a robotic arm for curtain wall installation by employing a sixth-order B-spline interpolation to ensure continuous motion trajectories. Secondly, 
We improved the NSGA-III algorithm by introducing a focused operator, resulting in an NSGA-III-FO algorithm. Through multiple comparative experiments on the DTLZ3 and WFG3 test functions, we verified that the NSGA-III-FO algorithm significantly enhances the convergence efficiency in multi-objective optimization problems. 
In a real-world curtain wall installation scenario, we designed two practical tasks and used the NSGA-III-FO algorithm to solve robotic arm trajectories that satisfy multi-objective constraints. Trajectory tracking experiments confirmed high consistency between the actual and pre-simulated trajectories. The experimental results demonstrate that the NSGA-III-FO algorithm is both efficient and practical in addressing multi-objective trajectory planning for curtain wall installation robots, providing new technical pathways and support for the development of construction robotics.
\addtolength{\textheight}{-12cm}   




\section*{ACKNOWLEDGMENT}

This work was supported in part by the National Key R\&D Program of China (No.2023YFB4705002), in part by the National Natural Science Foundation of China(U20A20283), in part by the Guangdong Provincial Key Laboratory of Construction Robotics and Intelligent Construction (2022KSYS 013), in part by the CAS Science and Technology Service Network Plan (STS) - Dongguan Special Project (Grant No. 20211600200062), in part by the Science and Technology Cooperation Project of Chinese Academy of Sciences in Hubei Province Construction 2023.

\end{document}